\documentclass[conference]{IEEEtran}
\IEEEoverridecommandlockouts
\usepackage{cite}
\usepackage{amsmath,amssymb,amsfonts}
\usepackage{algorithmic}
\usepackage{graphicx}
\usepackage{textcomp}
\usepackage{hyperref}
\usepackage{cleveref}
\usepackage{xcolor}
\usepackage{array}
\usepackage{siunitx}
\usepackage{booktabs} 
\usepackage{subcaption}
\usepackage{multirow} 
\usepackage[accsupp]{axessibility}  
\def\BibTeX{{\rm B\kern-.05em{\sc i\kern-.025em b}\kern-.08em
		T\kern-.1667em\lower.7ex\hbox{E}\kern-.125emX}}
\begin{document}

\author{
	\IEEEauthorblockN{
		Jiaqi Zhang\textsuperscript{1*},
		Zhuodong Liu\textsuperscript{2*},
		Kejian Yu\textsuperscript{3$\dagger$}
	}
	\IEEEauthorblockA{\textsuperscript{1} College of Computer and Information Science, Southwest University, Chongqing 400700, China}
	\IEEEauthorblockA{\textsuperscript{2} School of Economics and Management, Beijing Jiaotong University, Beijing 100044, China}
	\IEEEauthorblockA{\textsuperscript{3} School of Computer Science and Technology, Donghua University, Shanghai 201620, China}
	\thanks{\textsuperscript{*} These authors contributed equally to this work.}
	\thanks{\textsuperscript{$\dagger$} Corresponding author: Kejian Yu (yukejian2021@outlook.com)}
}
	\title{MSFNet-CPD: Multi-Scale Cross-Modal Fusion Network for Crop Pest Detection} 
	
\maketitle
\begin{abstract}
	Accurate identification of agricultural pests is essential for crop protection but remains challenging due to the large intra-class variance and fine-grained differences among pest species. While deep learning has advanced pest detection, most existing approaches rely solely on low-level visual features and lack effective multi-modal integration, leading to limited accuracy and poor interpretability. Moreover, the scarcity of high-quality multi-modal agricultural datasets further restricts progress in this field.
	To address these issues, we construct two novel multi-modal benchmarks—CTIP102 and STIP102—based on the widely-used IP102 dataset, and introduce a Multi-scale Cross-Modal Fusion Network (MSFNet-CPD) for robust pest detection. Our approach enhances visual quality via a super-resolution reconstruction module, and feeds both the original and reconstructed images into the network to improve clarity and detection performance. To better exploit semantic cues, we propose an Image-Text Fusion (ITF) module for joint modeling of visual and textual features, and an Image-Text Converter (ITC) that reconstructs fine-grained details across multiple scales to handle challenging backgrounds. Furthermore, we introduce an Arbitrary Combination Image Enhancement (ACIE) strategy to generate a more complex and diverse pest detection dataset, MTIP102, improving the model's generalization to real-world scenarios.
	Extensive experiments demonstrate that MSFNet-CPD consistently outperforms state-of-the-art methods on multiple pest detection benchmarks. All code and datasets will be made publicly available at: \url{https://github.com/Healer-ML/MSFNet-CPD}.
	
\end{abstract}
\section{Introduction}
Over the years, crop pests have consistently threatened agricultural yields, product quality, and profitability. Traditional methods of pest identification are slow, costly, and heavily reliant on expert knowledge\cite{a1}. However, recent advancements in deep learning and computer vision have led to the development of innovative pest classification techniques, such as Convolutional Neural Networks (CNNs) and Transformers. These techniques provide faster, more accurate, and cost-effective solutions to the challenges of pest detection.
\par In recent years, various deep learning techniques have been applied to crop pest detection to enhance accuracy and efficiency. You et al.\cite{a2} developed an offline mobile app that uses a compressed CNN to diagnose citrus pests, which stands out for its low cost, speed, and accuracy. Similarly, Wang et al.\cite{a3} designed a CNN with an Inception module to identify multiple crop diseases, achieving over 95\% accuracy. Their model is not only faster but also more precise compared to traditional methods. H.T. et al.\cite{a4} enhanced CNNs by incorporating attention mechanisms and feature pyramids, achieving 99.78\%  accuracy on a small pest dataset and yielding respectable results on the larger IP102 dataset. Guo et al.\cite{a5} advanced multi-label pest classification by applying the Swin Transformer, which significantly improved accuracy on the IP102 dataset. Wang et al.\cite{a6} developed a deep learning ensemble that combines CNNs and Swin Transformers, demonstrating high accuracy across various datasets. Bao et al.\cite{a7} employed DenseNet with a Coordinated Attention mechanism to accurately classify the severity of cotton aphid damage. Nigam et al.\cite{a8} fine-tuned EfficientNet for identifying wheat diseases, achieving high accuracy. Lastly, Yu et al.\cite{a9} combined Transformers with convolutional networks to create the Inception Convolution Visual Transformer (ICVT), which enhances feature extraction for plant disease detection through Dynamic Pattern Decomposition, resulting in improved performance in deep learning classifiers and machine learning models\cite{a10}.
\par The challenge of pest image detection arises from the complexity and variability of agricultural backgrounds, coupled with the scarcity of high-quality pest images. Historically, studies have focused on leaf-based pest detection. However, advancements in smart farming have highlighted the importance of integrating deep learning into this field \cite{a11}, despite the challenges posed by diverse agricultural data sources and the need for semantic interpretation \cite{a12}. Rice fusion Multi-modal approaches are being explored to fuse different types of data. For instance, Rutuja et al. \cite{a13} introduced rice fusion for rice disease diagnosis. In contrast, Zhang et al. \cite{a14} developed the MMFGT model, which utilizes self-supervised learning for fine-grained pest detection. Zhou et al. \cite{a15} investigated semantic embedding for disease image-text correlation, and Zhang et al. \cite{a16} implemented the Multi-ResNet34 method for diagnosing tomato diseases, contributing to pest detection technology advancements.
\par Although significant progress has been made in crop pest detection, several challenges remain. First, the complexity and diversity of agricultural environments can negatively impact the accuracy of models that primarily rely on low-level image features. Second, many models are designed for single-target detection and struggle to handle images with multiple pests of varying species and sizes. Finally, the performance of these models is often limited by the low quality of images captured in natural settings in the real world. To address these challenges, this study proposes a multi-scale cross-modal fusion network to enhance pest detection performance in the IP102 dataset. We constructed two multi-modal datasets, CTIP102 and STIP102, by creating simple and complex text descriptions for each pest image to combine visual and text features for multi-modal learning. We employ the Enhanced Super-Resolution Generative Adversarial Network (ESRGAN) \cite{a17} to perform super-resolution reconstruction on low-resolution images, preserving and enriching the pest feature information. Both the original and reconstructed images are used together for model training. Additionally, we introduce the ACIE data augmentation algorithm to improve the model's robustness and adaptability in real-world environments by generating a multi-target detection dataset, MTIP102. Our contributions to this study can be summarized as follows:
\par(1) This paper proposes the Multi-Scale Cross-modal Fusion Network (MSFNet-CPD) for crop pest detection. This approach is the first to integrate image high-frequency information with text for pest detection by recovering high-frequency information from low-quality images and combining visual and text features.
\par(2) Based on the IP102 dataset, we created two multi-modal datasets, STIP102 and CTIP102. Additionally, we introduced a generalized data enhancement algorithm, ACIE, to create the multi-target detection dataset MTIP102, which enhances the model's practical applicability.
\par(3) Extensive experiments on the constructed datasets validate the model's effectiveness, and comparisons with other models demonstrate the advantages of our approach.
\section{Method}
This section describes each part of the MSFNet-CPD model in detail, as shown in Figure \ref{img1}.
\subsection{Low and Super Resolution Generative Adversarial Network (LSRGAN)}
\begin{figure*}[htp]
	\centering
	\includegraphics[height=9cm]{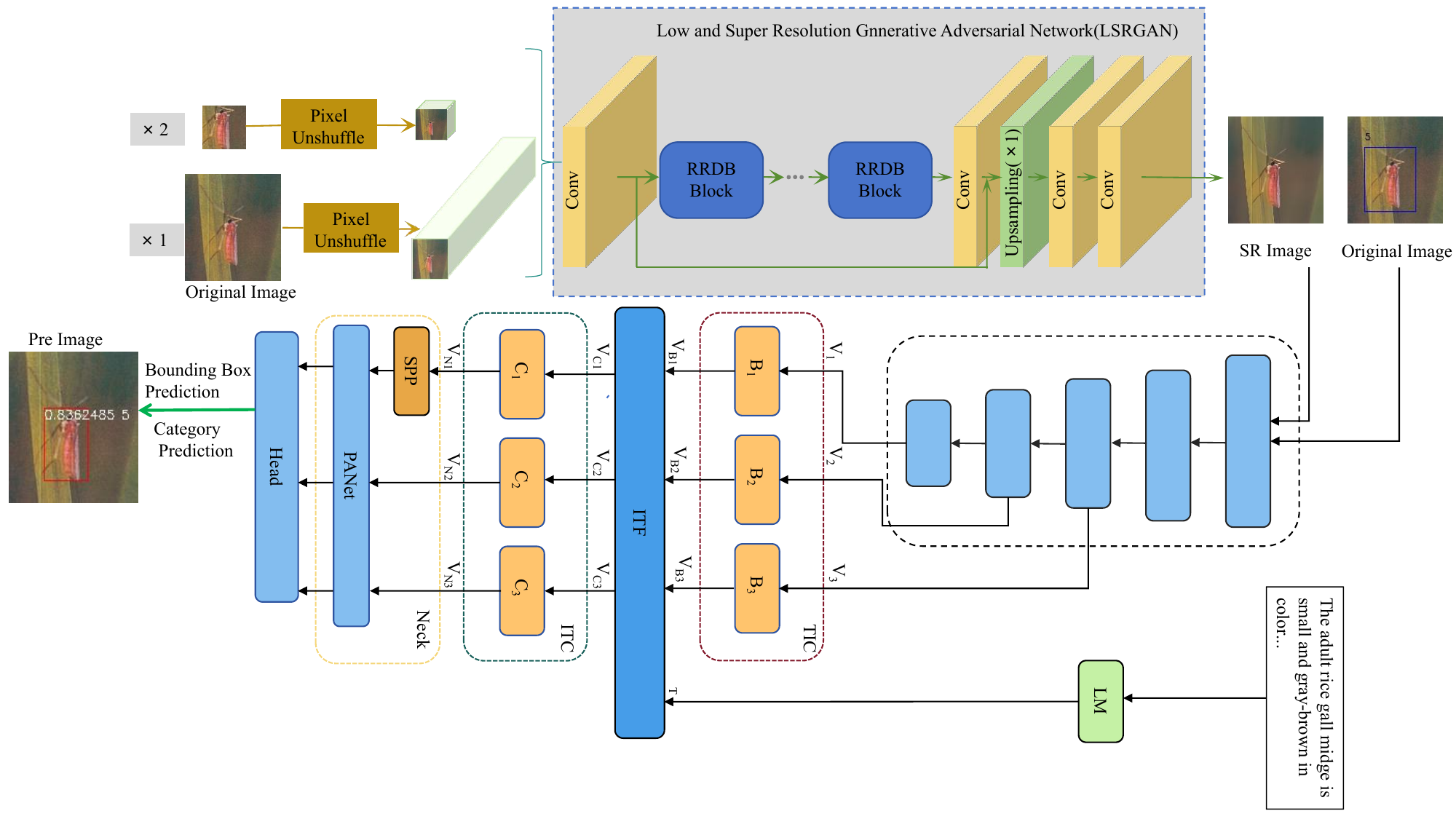}
	\caption{MSFNet-CPD Model Architecture (A. LSRGAN for super-resolution of low-quality images, B. Picture to ITF Transformer (TIC), C. Multi-scale Cross-modal Fusion ITF Module, D. ITF to Neck Network Converter (ITC), and E. Pest Target Identification (PTI). Numbers represent categories in Original Image and Pre Image, numbers represent class confidence and category).}
	\label{img1}
\end{figure*}
The low and super-resolution modules mainly convert low-resolution input data into high-resolution image formats. Low-resolution images may be due to small spatial resolution of image data or distortion such as blurring in the image. In addition, image texture loss, mutation loss, content loss, and pixel loss can be detected and repaired using the super-resolution (SR) method\cite{a18}. However, since some of the low-resolution information will be changed when reconstruction is performed, in this paper, the original image is used as a supplement to the SR image features and sent to the model for feature extraction, as shown in the upper part of Figure \ref{img1}.
\subsection{Text-Image Converter (TIC)}
In this paper, we design a multi-scale feature extraction block using a convolutional neural network. It processes the original and super-resolution images to extract features at three scales, which are then concatenated. The TIC aligns these visual features with text dimensions and reduces computational load. TIC consists of three parts ($B_1$, $B_2$, $B_3$), corresponding to visual features ($V_1$, $V_2$, $V_3$), and each part includes a convolutional layer, ReLU layer, and max pooling layer. Input dimensions for $B_1$, $B_2$, and $B_3$ are 19×19×1024, 38×38×512, and 76×76×256, respectively. The output was scaled to 5 × 5 × T and then reshaped to 25 visual feature markers for ITF input, with T representing the text size, as shown in the specific module parameters in Table \ref{tab:1}.
\subsection{Multi-Scale Cross-Modal Fusion ITF Module }
\begin{figure}[htbp]
	\centering
	\includegraphics[height=4cm]{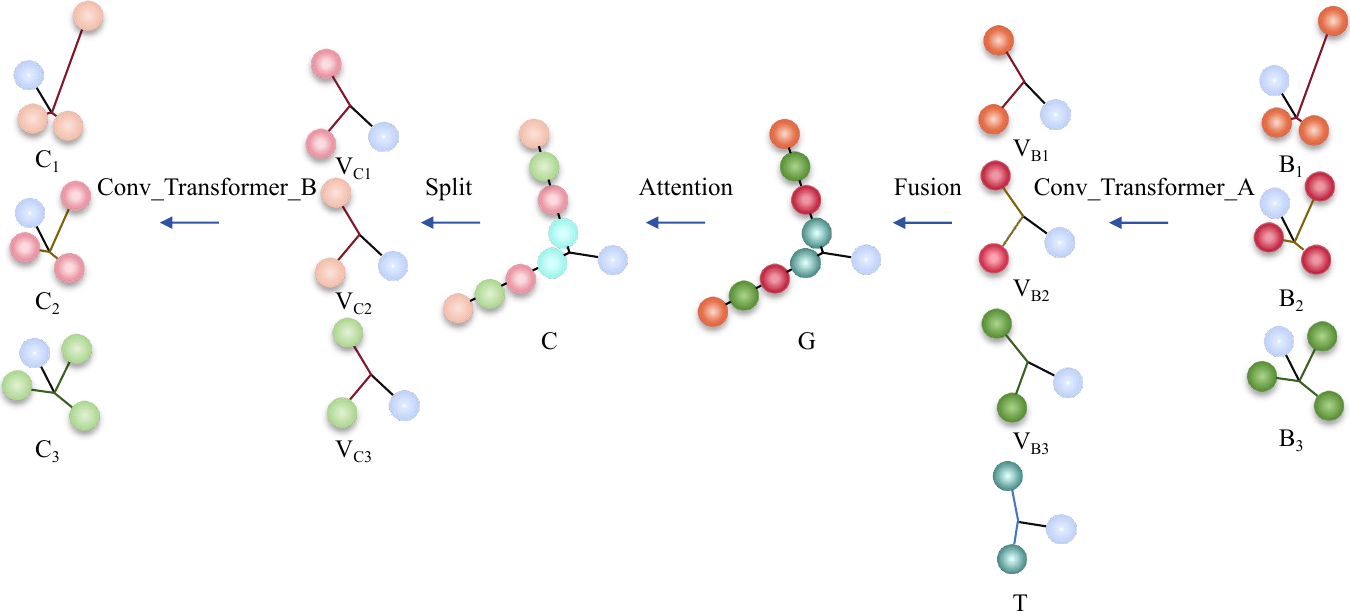}
	\caption{ITF Specific Process. ($B_i$, $C_i$, $V_i$ stand for different scale image features and $T$ stands for text features).}
	\label{img2}
\end{figure}

ITF, based on the Transformer encoding architecture proposed by Vaswani et al.\cite{a19}, is demonstrated in Figure \ref{img2} to illustrate the specific internal process of ITF. Conv\_Transformer\_A extracts visual features and transforms them into the Transformer architecture encoding through convolution. This structure includes multi-head self-attention sub-layers and fully connected feed-forward sub-layers. 
Residual connections and layer normalization are applied between the two sub-layers. The Transformer encoder employs scaled dot-product attention, defined as follows: 
\begin{equation}
	Attention(Q,K,V) = soft\max (\frac{{Q{K^T}}}{{\sqrt {{{\rm{d}}_{\rm{k}}}} }}){\rm{V}}
	\label{eq:1}
\end{equation}
$Q$, $K$, and $V$ consist of queries, keys, and values, respectively, and $d_k$ represents the dimensionality of keys. In our model, we concatenate text and image features into a new sequence $G$:
\begin{equation}
	\scalebox{0.7}{$
		{{G}_{LR}, {G}_{SR}}{\rm{ = }} 
		\underbrace {{t_1},{t_2}, \cdot  \cdot  \cdot ,{t_n}}_n, 
		\underbrace {V_{B1}^{1,1}, \cdot  \cdot  \cdot ,V_{B1}^{5,5}}_{25}, 
		\underbrace {V_{B2}^{1,1}, \cdot  \cdot  \cdot ,V_{B2}^{5,5}}_{25}, 
		\underbrace {V_{B3}^{1,1}, \cdot  \cdot  \cdot ,V_{B3}^{5,5}}_{25}
		$}
	\label{eq:2}
\end{equation}

\begin{equation}
	{{G}} = {\rm{Concat}}({{{G}}_{{{LR}}}},{{{G}}_{{{SR}}})}
	\label{eq:3}
\end{equation}
${G}_{LR}$, ${G}_{SR}$ represent high-resolution image features and low-resolution image features respectively. The reconstructed image features before and after concatenation into $G$ are input into ITF; hence, $Q$=$K$=$V$=$G^T$. Ultimately, ITF outputs multi-scale image features $V_{C1}$, $V_{C2}$ and $V_{C3}$, which are encoded through Conv\_Transformer\_B for subsequent ITC. Conv\_Transformer\_B is the reverse process of Conv\_Transformer\_A.
\subsection{Image-Text Converter (ITC)}
The ITC is the inverse process of the TIC. The ITC outputs features of the same size ($V_{Ni}$) as the input features ($V_i$) to the TIC, where i $\in$ [1,3]. The ITC consists of three parts, denoted as $C_1$, $C_2$, and $C_3$ corresponding to $B_1$, $B_2$, and $B_3$ in the TIC, with specific architectures and parameters detailed in Table \ref{tab:1}. Each part comprises transpose convolution (ConvT)\cite{a20} ReLU and upsampling operations. ConvT performs the reverse operation of convolution. Since the kernel size, padding size, and stride of ConvT are the same as those used in the Conv layer, ConvT generates image features with the exact dimensions as the input features. Upsampling reverses the pooling operation using the nearest-neighbor interpolation algorithm.
\begin{table}[htp]
	\caption{TIC and ITC Structure.}
	\label{tab:1}
	\centering
	\scalebox{0.8}
	{
		\begin{tabular}{llll}
			\toprule
			\multirow{6}{*}{TIC} & \multirow{2}{*}{B1} & \multicolumn{1}{c} {Conv} & \multicolumn{1}{c}{[K=3x3, P=1, S=2]} \\ 
			& & MaxPooling & \multicolumn{1}{c}{[2x2]} \\ 
			& \multirow{2}{*}{B2} & \multicolumn{1}{c} {Conv} & \multicolumn{1}{c}{[K=3x3, P=1, S=2; K=3x3, P=1, S=2]} \\ 
			& & MaxPooling & \multicolumn{1}{c}{[2x2]} \\ 
			& \multirow{2}{*}{B3} & \multicolumn{1}{c} {Conv} & \multicolumn{1}{c}{[K=5x5, P=2, S=4; K=3x3, P=1, S=2]} \\ 
			& & MaxPooling & \multicolumn{1}{c}{[3x3]} \\ 
			\midrule
			\multirow{6}{*}{ITC} & \multirow{2}{*}{C1} & UpSampling & \multicolumn{1}{c}{[2x2]} \\ 
			& & \multicolumn{1}{c} {ConvT} & \multicolumn{1}{c}{[K=3x3, P=1, S=2]} \\ 
			& \multirow{2}{*}{C2} & UpSampling & \multicolumn{1}{c}{[2x2]} \\ 
			& & \multicolumn{1}{c} {ConvT} & \multicolumn{1}{c}{[K=3x3, P=1, S=2; K=3x3, P=1, S=2]} \\ 
			& \multirow{2}{*}{C3} & UpSampling & \multicolumn{1}{c}{[2x2]} \\ 
			& & \multicolumn{1}{c} {ConvT} & \multicolumn{1}{c}{[K=3x3, P=1, S=2; K=5x5, P=2, S=4]} \\  
			\bottomrule
	\end{tabular}}
	\caption*{$B_1$, $B_2$, $B_3$, $C_1$, $C_2$ and $C_3$ stand for the corresponding module name of Figure \ref{img1}. K, S, P are the convolution kernel, step size, and padding respectively.}
\end{table}
\subsection{Pest Target Identification (PTI) }
This paper primarily compares the advantages and disadvantages of unimodal and multi-modal approaches. Due to its simpler structure and fewer model parameters than YOLOv8 and YOLOv9, we utilize the Neck and Head Networks of YOLOv4 \cite{a21} for pest target detection. The multi-scale cross-modal features $V_{N1}$, $V_{N2}$  and  $V_{N3}$ from the image feature extraction network are connected to the neck network, which includes Spatial Pyramid Pooling (SPP) \cite{a22} and Path Aggregation Network (PANet) \cite{a23}. The SPP consists of three maximum pooling layers, while the PANet comprises convolutional and downsampling components. PANet enhances the receptive field and retains spatial information. The processed information at different scales is then fed into the head network, which predicts three types of bounding boxes at different scales. Detailed calculations for bounding offsets can be found in Redmon and Farhadi \cite{a24}. In our model, num\_class is set to 102 to account for the 102 different pest classes. The PTI objective function includes the bounding box regression loss ${\cal L}_{\rm{B}}$ and the target class loss ${\cal L}_{\rm{O}}$:
\begin{equation}
	{{\cal L}_{PTI}} = {{\cal L}_{\rm{B}}} + {{\cal L}_{\rm{O}}}
	\label{eq:4}
\end{equation}
As with YOLOv4, the bounding box regression loss of  ${\cal L}_{\rm{B}}$ includes the Mean Squared Error (MSE) of the width and height of the predicted box versus the width and height of the target box. Additionally, ${\cal L}_{\rm{O}}$ is the target category loss, which calculates the binary cross-entropy (BCE) of the predictor box category probability with the category information of the target box. 
\subsection{Data Set Construction }
In this study, we utilize two types of data: image and text. The image data are mainly from the IP102\cite{a25} public dataset, which contains photos of forest and field pests of variable quality and often affected by noise. To ensure the quality of the data, we manually screened and extracted some high-quality images for model comparison experiments, and this dataset was named HIP102. To compensate for the lack of textual feature information, we constructed two text descriptions, simple and complex, for each pest. These text descriptions are intended to provide information about the pest characteristics and thus increase the source of information for the model. 
\par
\begin{figure}[t]
	\centering
	\includegraphics[height=4cm]{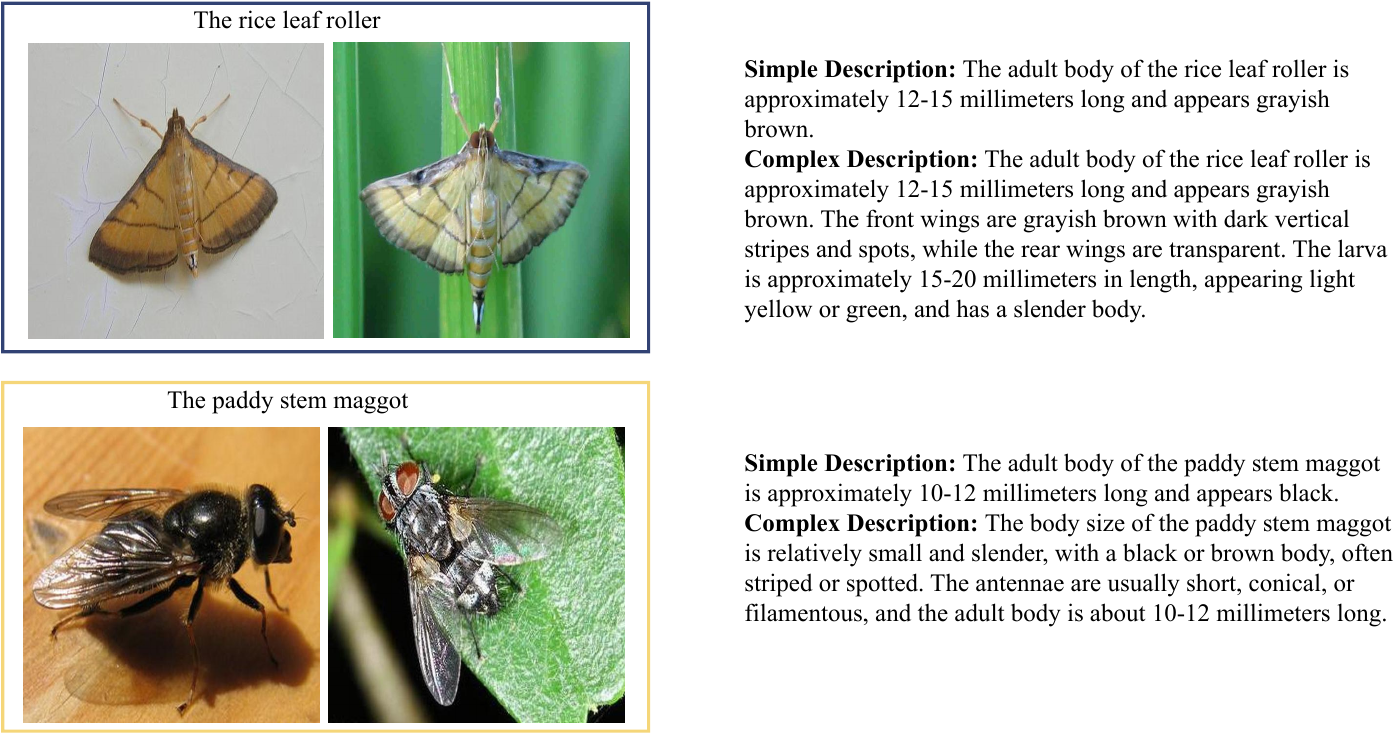}
	\caption{Partial Presentation of Multi-modal Dataset.}
	\label{img4}
\end{figure}

 In the process of text data collection, to ensure the credibility of the data. Our text data were mainly obtained from specialized pest books. In addition, we also applied crawler technology to obtain relevant text data from authoritative websites. These text descriptions mainly cover the main features of pests, such as morphology, size, color, etc., as shown in Figure \ref{img4}. 
\subsection{Arbitrary Combination Image Enhancement }
In real agricultural environments, pests live in sophisticated and diverse conditions, with multiple pests often appearing on the same crop simultaneously. Photos taken at different angles can result in pest size and relative position variations. Most current datasets contain images with only 1 or 2 pests, which are relatively fixed in size and have high overlap, limiting pest diversity in the training data and affecting model prediction accuracy. To address these issues, this study proposes an image data enhancement algorithm. Unlike traditional methods such as image rotation, noise addition, and saturation adjustment, this method fully considers pest diversity and environmental randomness. It reduces labeling costs and better adapts the model to identify pests in real scenarios. The specific steps of the algorithm are as follows:
\begin{table}[htp]
	\centering
	\begin{tabular}{@{}l@{}}
		\textbf{Algorithm ACIE}  \\
		\hline
		Input: Target Images($T$); Background Images($B$); The number of targets $R$ \\on the 
		image and the number of generated images $num$ \\
		Output: $num$ images containing $R$ target images \\
		\hline
		1: Repeat \\
		2: \hspace{0.5em}for \_ in range($num$) \\
		3: \hspace{0.5em} Select $B$ \\
		4: \hspace{2em} for $r\leftarrow 1$ to $R$ \\
		5: \hspace{3em} Select $T$ Calculate $w, h$\\ 
		\hspace{4.2em}(minimum bounding rectangle width and height).\\
		\hspace{4.2em} Randomly select a point $(x_1, y_1)$, place target $T$ on background $B$, \\
		\hspace{4.2em} and calculate the final position based on $w, h$ \\
		6: \hspace{3em}  Obtain image annotations $[(x_1, y_1), (x_2, y_2), {{class\_id}}]$ \\
		7: \hspace{3em}  If Target overlap \\
		8: \hspace{5em}  Go to step 6 \\
		9: \hspace{2em}  Until Target does not overlap \\
		10: \hspace{0.5em} Return Composite image \\
		\hline
	\end{tabular}
\end{table}
\begin{figure}[htbp]
	\centering
	\subcaptionbox{Simple}{
		\includegraphics[width=0.45\linewidth]{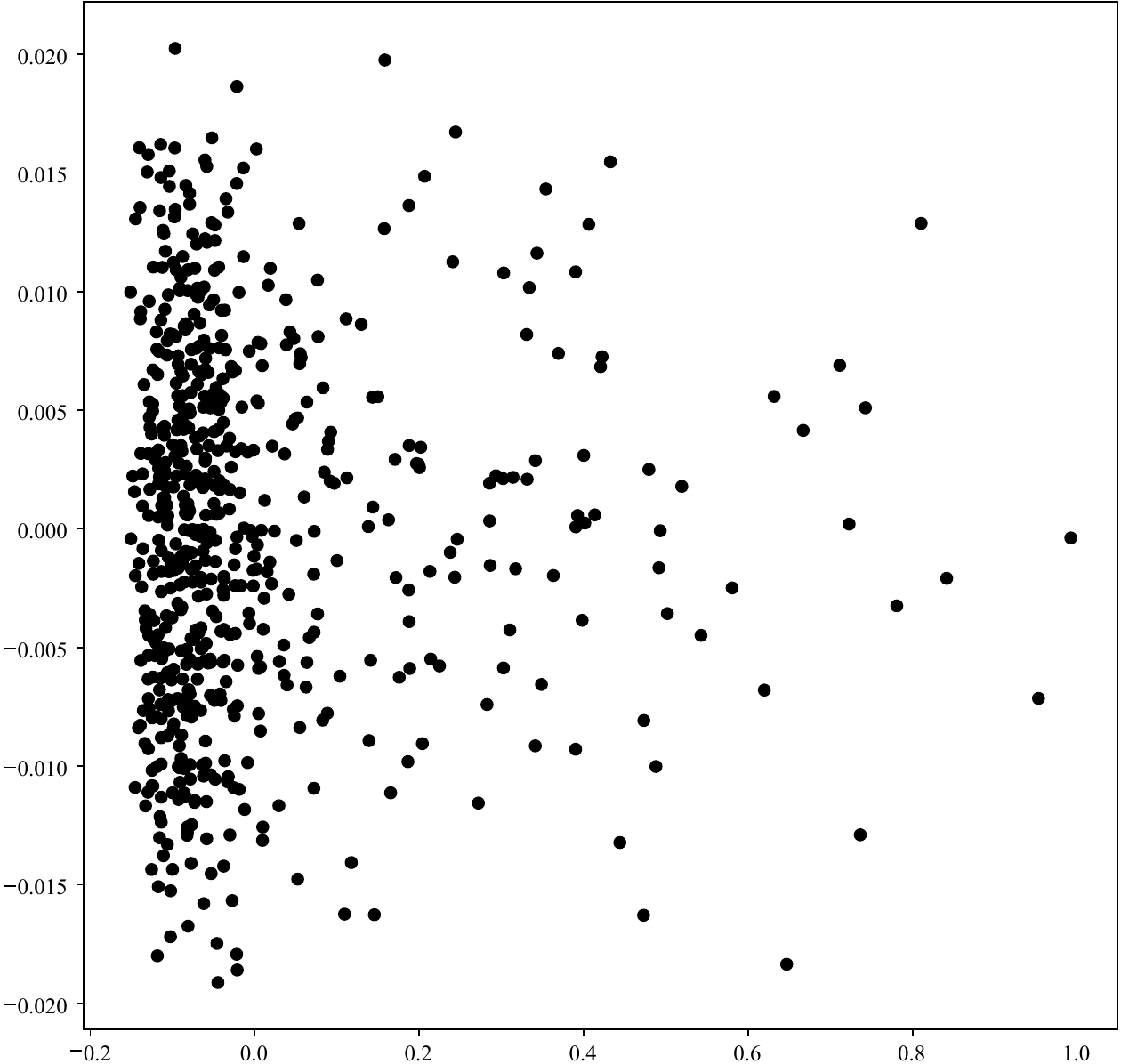}}
	\hspace{0.01\linewidth}
	\subcaptionbox{Complex}{
		\includegraphics[width=0.45\linewidth]{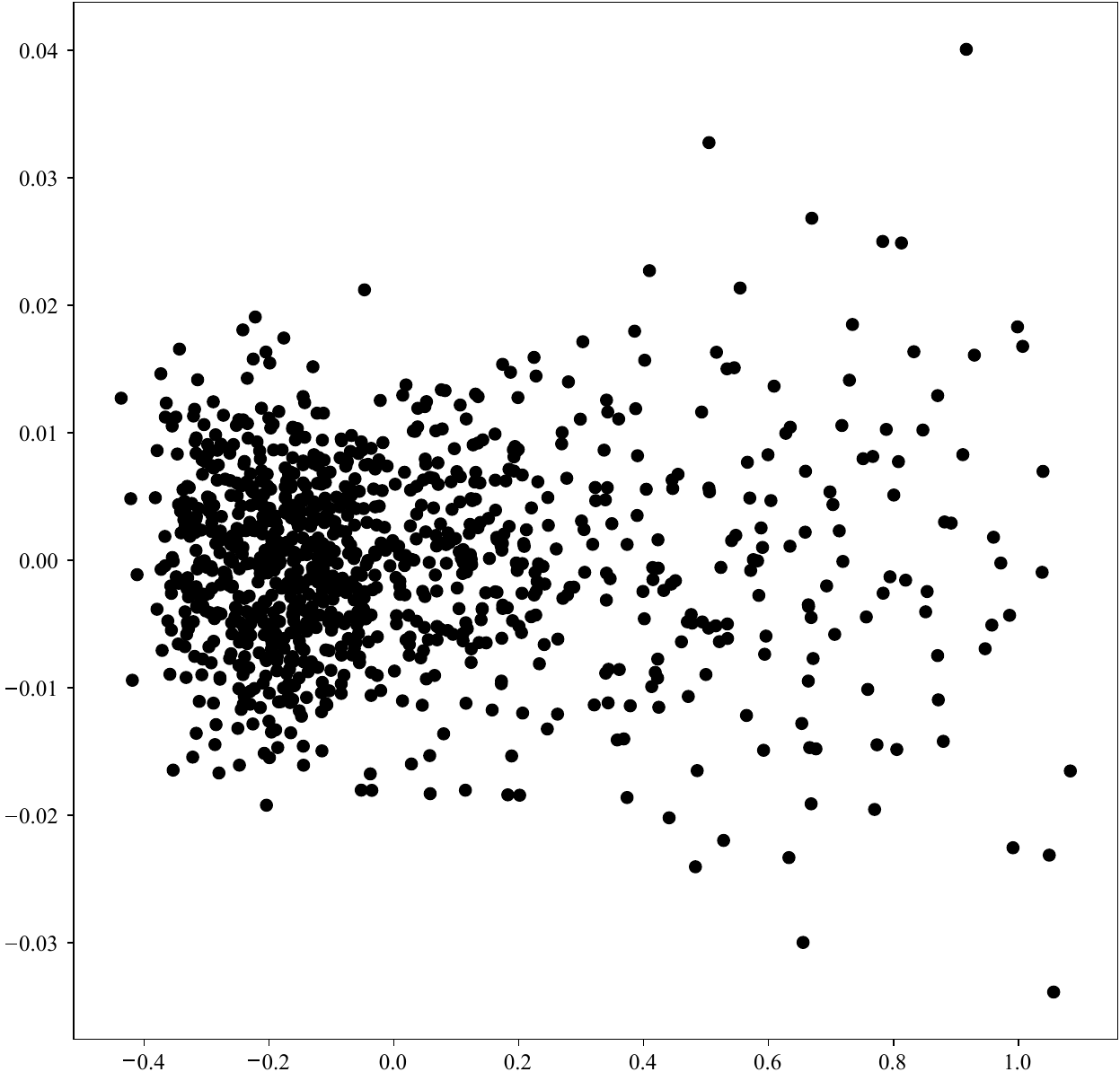}}
	\vfill
	\subcaptionbox{Simple}{
		\includegraphics[width=0.45\linewidth]{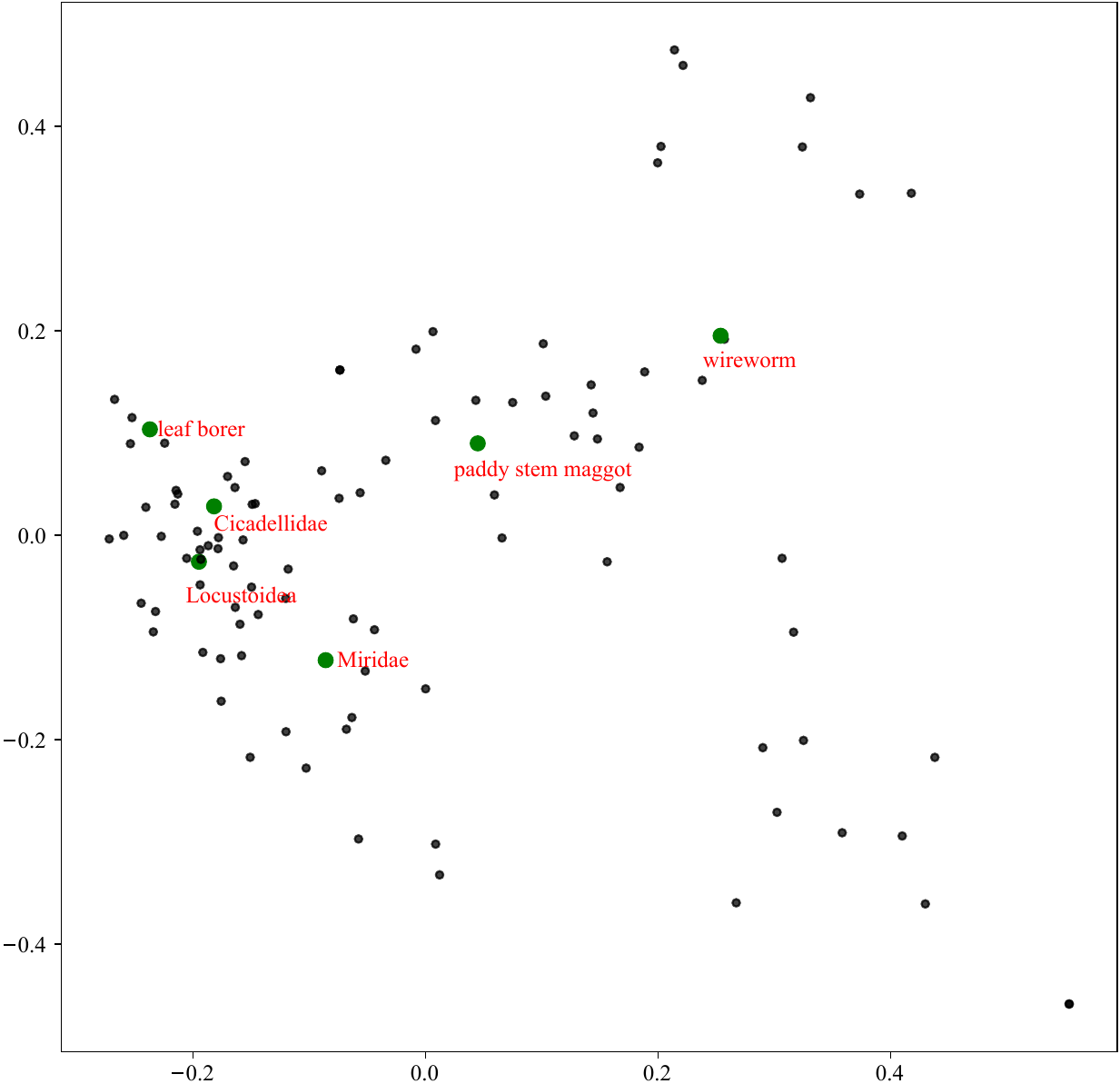}}
	\hspace{0.01\linewidth}
	\subcaptionbox{Complex}{
		\includegraphics[width=0.45\linewidth]{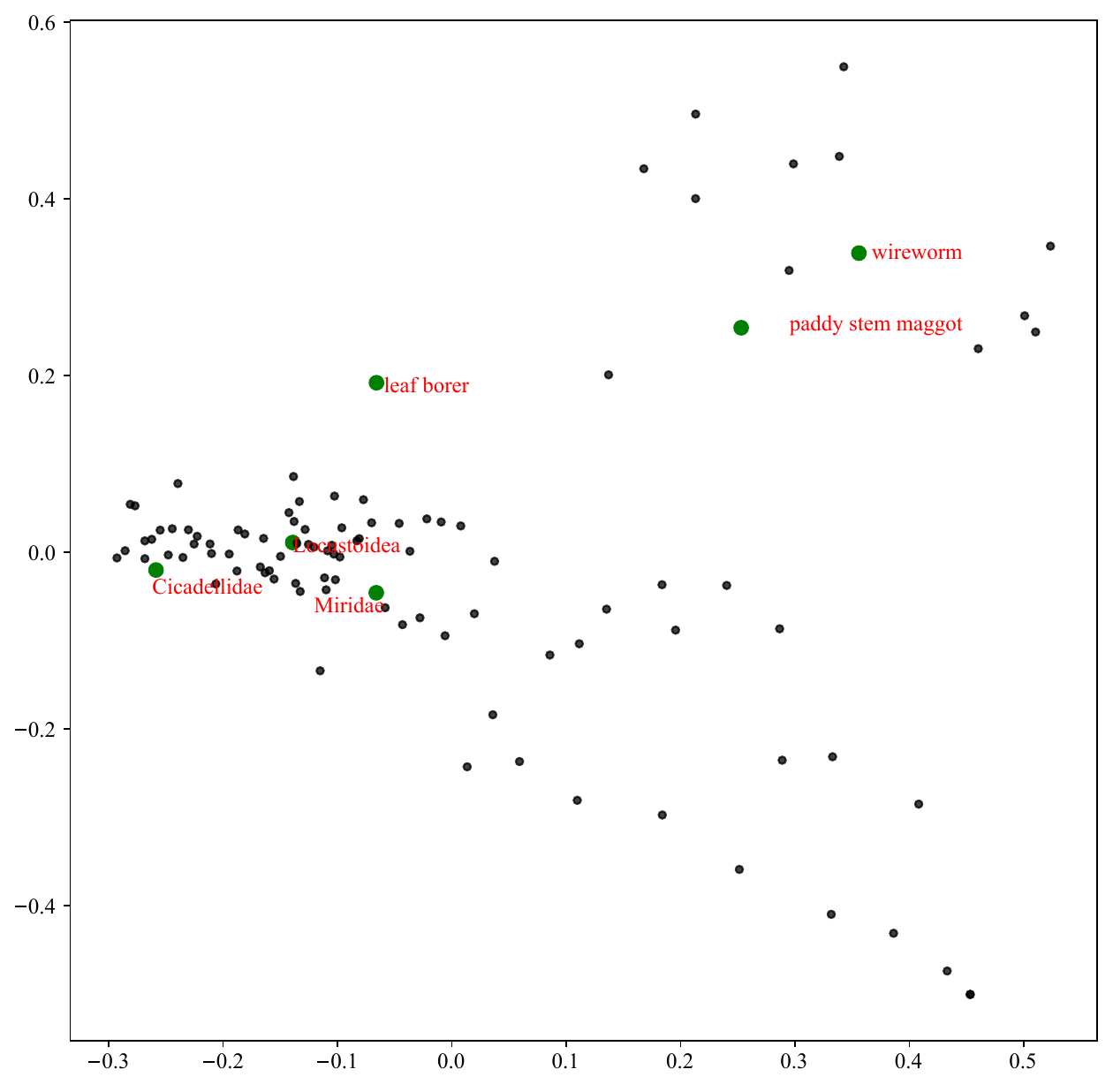}}
	\caption{Visualization of the semantic correlation analysis of simple and complex text descriptions. (Horizontal and vertical coordinates in the figure indicate the coordinates after mapping the feature vectors into a low-dimensional space).}
	\label{img9}
\end{figure}
\par In the above ACIE, $B$, $T$, $R$, and $num$ in this study are 580, 820, 4, and 10,000, respectively. This innovative data enhancement method effectively increases the pest diversity and randomness in the training dataset, provides more challenging training data for the model, and boosts the model's performance for application in real environments. 
\section{Experimentation }
\subsection{Data Setup }
In this study, we applied five datasets, IP102, HIP102, STIP102, CTIP102, and MTIP102, dividing the datasets according to 8:1:1. Table \ref{tab:2} shows the number of images and text characters in the training, validation, and test sets of these datasets. We use these datasets to evaluate the performance of our proposed MSFNet-CPD model.
\begin{table*}[ht]
	\centering
	\caption{Dataset Description and Division.}
	\label{tab:2}
	\scalebox{1.2}
	{
		\begin{tabular}{>{\centering\arraybackslash}m{2cm} >{\centering\arraybackslash}m{2cm} >{\centering\arraybackslash}m{2cm} >{\centering\arraybackslash}m{2cm} >{\centering\arraybackslash}m{2cm} >{\centering\arraybackslash}m{2cm}}
			\toprule
			\multicolumn{2}{c}{Dataset}  & Training & Test & Validation & Total \\
			\midrule
			\multirow{2}{*}{IP102} & Images & 60178 & 7522 & 7522 & 75222 \\ 
			& Text & $-$ & $-$& $-$& $-$ \\
			\multirow{2}{*}{HIP102} & Images & 15180 & 1898 & 1898 & 18976 \\ 
			& Text & 306124 & 42513 & 43214 & 391851 \\
			\multirow{2}{*}{STIP102} & Images & 60178 & 7522 & 7522 & 75222 \\ 
			& Text & 12626878 & 405124 & 405080 & 13437082 \\
			\multirow{2}{*}{CTIP102} & Images & 60178 & 7522 & 7522 & 75222 \\ 
			& Text & 34501251 & 951324 & 952652 & 36405227 \\
			\multirow{2}{*}{MTIP102} & Images & 8000 & 1000 & 1000 & 10000 \\ 
			& Text & 665549 & 83083 & 83083 & 831715 \\
			\bottomrule
	\end{tabular}}
	\caption*{The IP102 dataset serves as the original baseline collection. HIP102 is a refined subset of hand-screened, high-quality images with detailed text descriptions. CTIP102 combines the original IP102 images with complex text descriptions, while STIP102 pairs the original images with more straightforward, concise text descriptions. MTIP102, constructed using the ACIE algorithm, includes images with complex text descriptions and is designed to enhance multiscale feature diversity for robust model training.}
\end{table*}
\begin{table}[ht]
	\centering
	\caption{Model Parameter Settings.}
	\label{tab:3}
	\scalebox{0.65}{
		\begin{tabular}{>{\centering\arraybackslash}m{4cm} >{\centering\arraybackslash}m{4cm} >{\centering\arraybackslash}m{4cm}}
			\toprule
			Experimental Settings & Hyperparameter & Optimized Value \\
			\midrule
			\multirow{6}{*}{Training Settings} & batch size & 4 \\ 
			& optimizer & Adam \\ 
			& weight decay & 0.0000001 \\ 
			& learning rate & 5.00E-05 \\ 
			& dropout & 0.5 \\ 
			& epochs & 30 \\
			\multirow{3}{*}{Image} & conf\_thresh & 0.5 \\ 
			& nms\_thresh & 0.4 \\
			& pretrain\_model & yolo \\
			\multirow{5}{*}{Text} & word\_maxlen & 41 \\ 
			& sent\_maxlen & 35 \\ 
			& hidden\_size & 768 \\ 
			& num\_attention\_heads & 16 \\ 
			& pretrain\_model & bert-base-uncased \\
			\bottomrule
		\end{tabular}
	}
\end{table}
\begin{table}[h]
	\centering
	\caption{IP102 Advanced Model Target Detection Effects.}
	\label{tab:4}
	\scalebox{1}	{
		\begin{tabular}{l l S[table-format=2.2] S[table-format=2.2] S[table-format=2.2]}
			\toprule
			Method & Backbone & {mAP(\%)} & {mAP$_{50}$(\%)} & {mAP$_{75}$(\%)} \\
			\midrule
			FRCNN\cite{a27} & VGG-16 & 21.05 & 47.87 & 15.23 \\ 
			FPN\cite{a28} & ResNet-50 & 28.10 & 54.93 & 23.30 \\ 
			SSD300\cite{a29} & VGG-16 & 21.49 & 47.21 & 16.57 \\ 
			RefineDet\cite{a30} & VGG-16 & 22.84 & 49.01 & 16.82 \\ 
			YOLOv3\cite{a24} & DarkNet-53 & 25.67 & 50.64 & 21.79 \\
			YOLOv8\cite{a36} & DarkNet & 38.41 & 73.58 & 29.17 \\
			YOLOv9\cite{a37} & E-ELAN & 42.32 & 81.10 & 39.86 \\
			\bottomrule
	\end{tabular}}
	\caption*{The results for FRCNN, FPN, SSD300, and RefineDet are from \cite{a25}, while the results for YOLOv8, YOLOv9, and YOLOv4 are from the experiments in this paper, all using the IP102 unimodal dataset.}
\end{table}
\begin{table}[h]
	\centering
	\caption{Indicator Results from Different Data Sets.}
	\label{tab:5}
	\scalebox{1}
	{
		\begin{tabular}{llllll}
			\toprule
			Dataset & P(\%) & F1(\%) & mAP(\%) & mAP$_{50}$(\%) & mAP$_{75}$(\%) \\
			\midrule
			HIP102 & 90.88 & 81.23 & 52.21 & 95.35 & 41.80 \\
			STIP102 & 72.43 & 69.15 & 44.98 & 88.22 & 42.30 \\
			CTIP102 & 82.15 & 78.21 & 46.06 & 92.18 & 40.18 \\
			MTIP102 & 45.24 & 51.23 & 22.33 & 43.04 & 20.78 \\
			\bottomrule
	\end{tabular}}
	\caption*{Experimental results of the MSFNet-CPD model using different multi-modal datasets in the same setup.}
\end{table}
\subsection{Experimental Setup } 
In this experiment, we implemented the algorithm on a Linux Ubuntu 18.04 workstation equipped with 70GB of RAM and an A6000-48G GPU using Pycharm, Python 3.7, and PyTorch 1.7.0+cu110. We used the Adam optimization algorithm for training and applied non-maximal suppression (NMS) for object detection. YOLOv4 and bert-base-uncased pre-trained models were used for image and text processing, with the specific parameter configurations listed in Table \ref{tab:3}. These settings validated the effectiveness of the MSFNet-CPD model for pest identification and classification.

\par In this study, we have used Word2Vec\cite{a34} and TF-IDF to visualize and analyze the semantic relevance of textual information, as shown in Figure \ref{img9}. Both methods show a high degree of aggregation of complex text descriptions at the word and sentence level. Thus, compared to STIP102, these results highlight the complexity and richness of text descriptions in the CTIP102 dataset, providing valuable insights for data processing and model training.

\subsection{Model evaluation and analysis}
We use several commonly adopted metrics to comprehensively evaluate model performance in object detection tasks: Precision, recall, F1-Score, and Average Precision (AP). Average Precision includes Mean Average Precision (mAP), calculated as the average over a range of Intersection over Union (IoU) values from 0.5 to 0.95, as well as mAP at specific thresholds: mAP$_{50}$ (IoU=0.5) and mAP$_{75}$ (IoU=0.75). These metrics are crucial for assessing the accuracy of pest classification and detection performance.
\par In this section, we evaluate the performance of our proposed MSFNet-CPD model using four multi-modal datasets derived from the IP102 dataset. The detailed results are shown in Table \ref{tab:5}. Additionally, Table \ref{tab:4} presents the mAP, mAP$_{50}$, and mAP$_{75}$ results from the unimodal IP102 dataset for the target detection task. The results indicate that the CTIP02 and STIP102 datasets outperform the unimodal dataset in detection performance. This suggests that incorporating textual descriptive features corresponding to the images helps overcome the limitations of using single-image features, thus improving the model’s extraction, detection, and classification effectiveness. This further demonstrates that the multi-modal dataset developed in this study is more complete and accurate than the unimodal IP102 dataset. We also explore the impact of text feature complexity on model performance. Experimental results from the STIP102 and CTIP102 datasets show that more complex text descriptions lead to better performance compared to simpler ones, especially during the feature extraction and fusion phases. Specifically, using complex text descriptions improves Precision, mAP, and mAP$_{50}$ by 9.72\%, 1.08\%, and 3.96\%, respectively, compared to more straightforward text descriptions. As shown in Table \ref{tab:4}, the mAP achieved using YOLOv9 is 42.32\%, with mAP$_{50}$ at 81.10\% and mAP$_{75}$ at 39.86\%. In comparison, our proposed model significantly outperforms this state-of-the-art model across all metrics—precision, mAP, mAP$_{50}$, and mAP$_{75}$. These results convincingly demonstrate that cross-modal feature learning is more effective than unimodal features in agricultural pest identification tasks, providing valuable insights for future research.
\begin{figure}[t]
	\centering
	\subcaptionbox{mAP$_{75}$\label{img11-1}}{
		\includegraphics[width=0.45\linewidth]{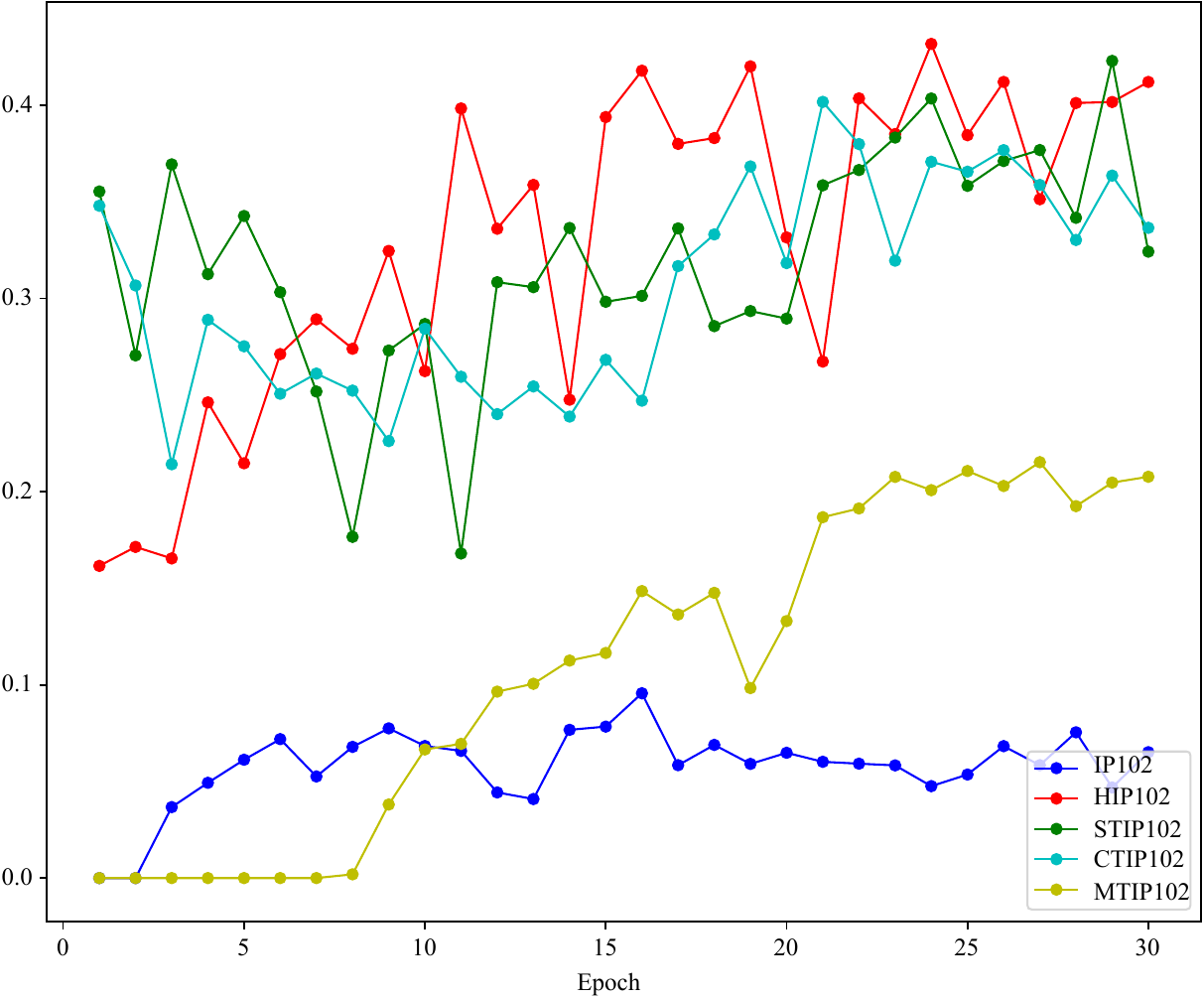}}
	\hspace{0.01\linewidth}
	\subcaptionbox{P\label{img11-2}}{
		\includegraphics[width=0.45\linewidth]{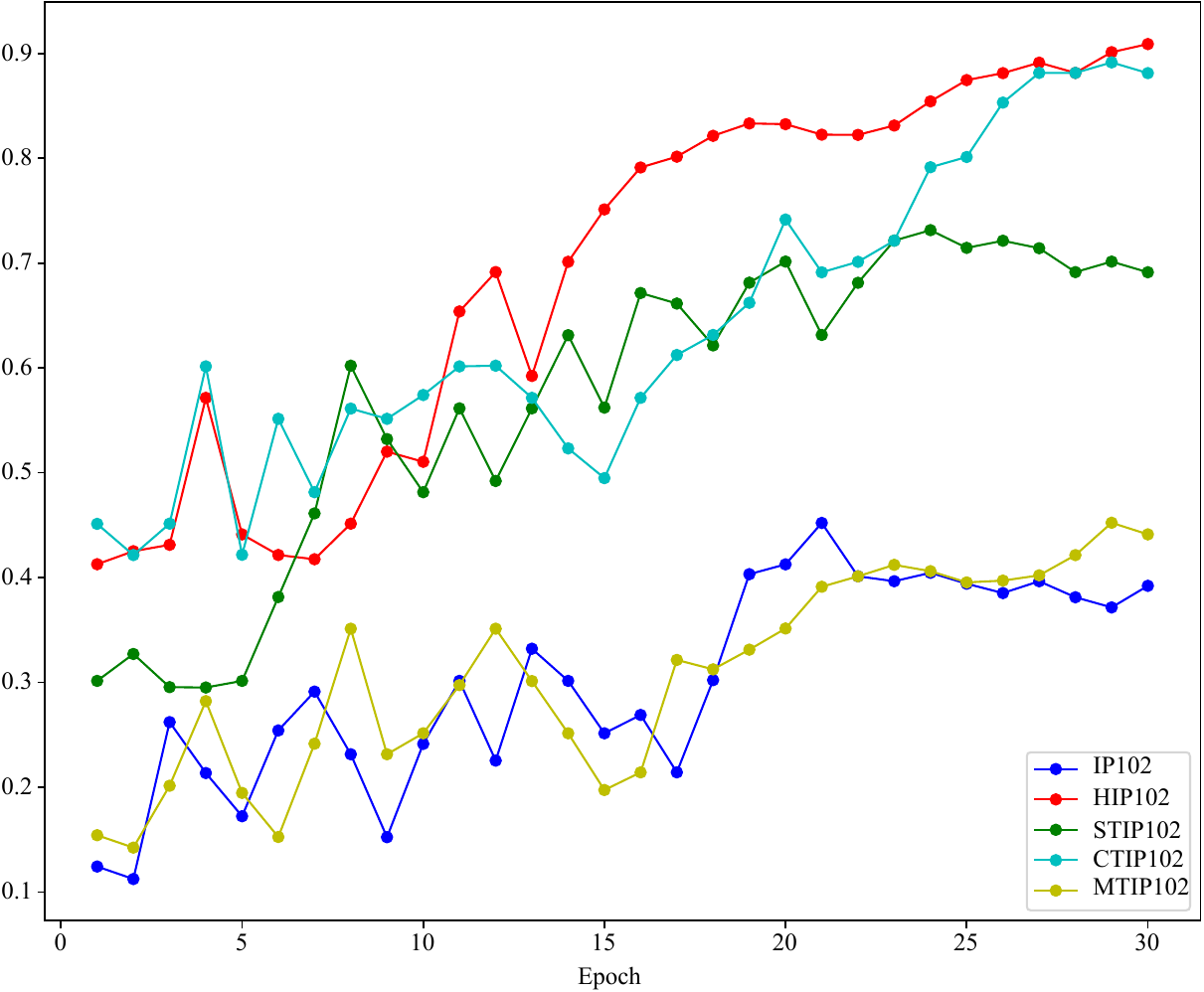}}
	\vfill
	\subcaptionbox{mAP\label{img11-3}}{
		\includegraphics[width=0.45\linewidth]{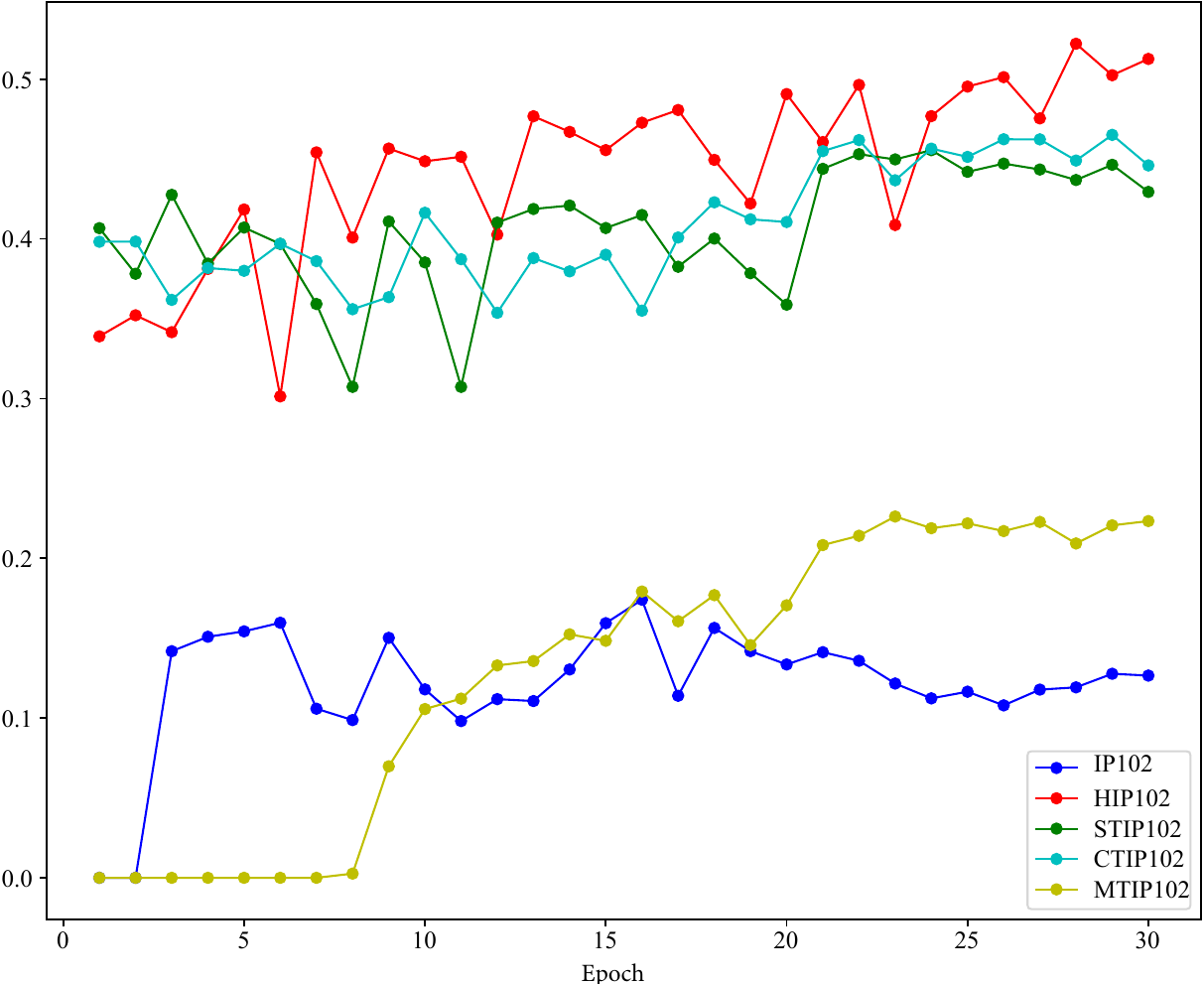}}
	\hspace{0.01\linewidth}
	\subcaptionbox{mAP$_{50}$\label{img11-4}}{
		\includegraphics[width=0.45\linewidth]{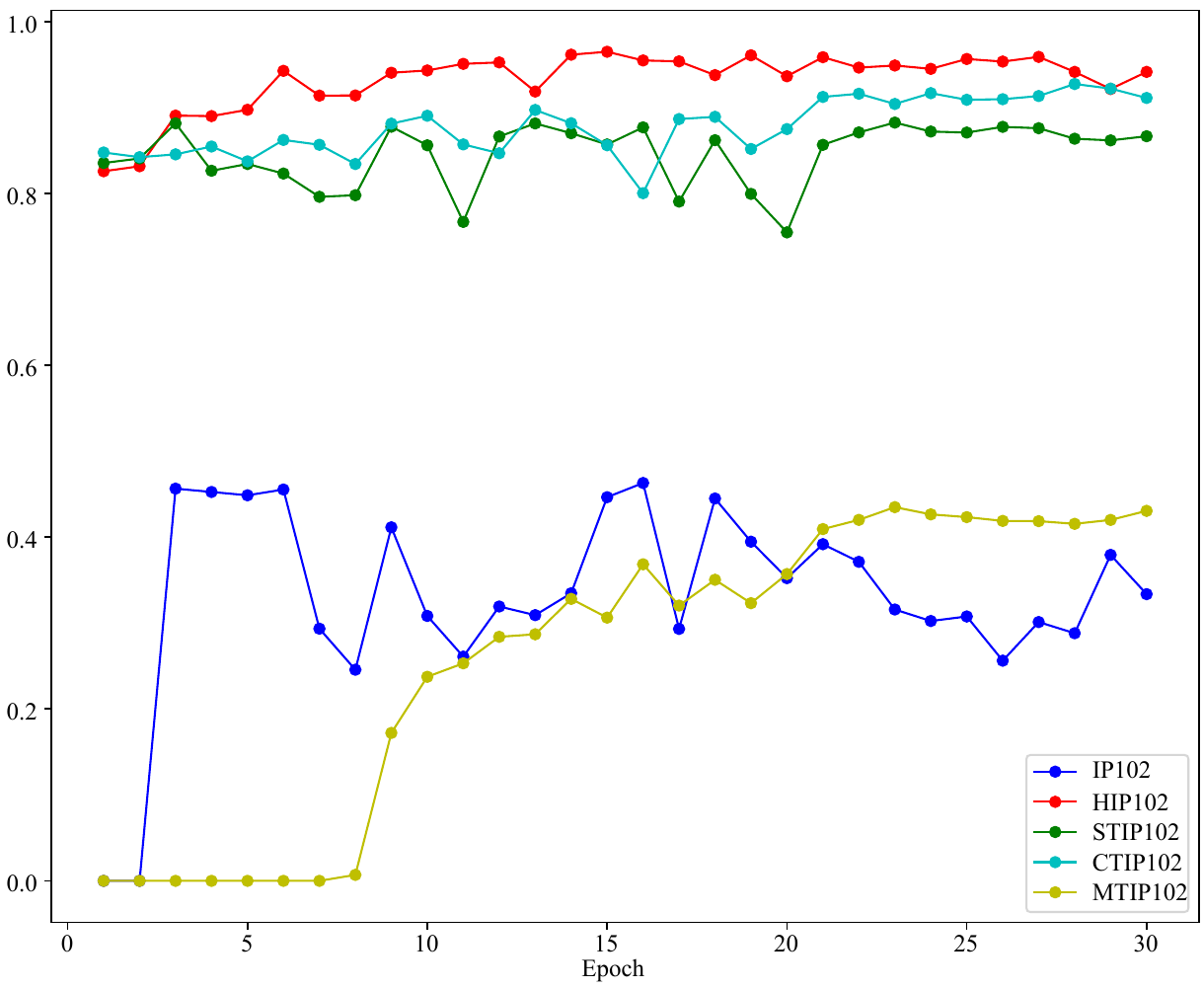}}
	\caption{Comparison of Indicators for Different Data Sets.}
	\label{img11}
\end{figure}

\subsection{Ablation Experiments }
Ablation experiments were performed by removing text (w/o text) and LSRGAN (w/o LSRGAN) from our multi-modal model. The results of these experiments are shown in Table \ref{tab:IP102}, where the backbone is CSPDakNet53, and the language model (LM) is BERTbase. When the text component was removed, the mAP decreased by 10.42\%, mAP$_{50}$ dropped by 10.5\%, and mAP$_{75}$ decreased by 10.76\%. Similarly, when the LSRGAN component was removed, mAP decreased by 11. 24\%, mAP$_{50}$ decreased by 12. 64\%, and mAP$_{75}$ decreased by 11.19\%.  These results highlight the importance of high-resolution image and text features for target detection.
\begin{table}[t]
	\centering
	\caption{Ablation Results.}
	\label{tab:IP102} 
	\scalebox{1}
	{\begin{tabular}{l S[table-format=2.2] S[table-format=2.2] S[table-format=2.2]}
			\toprule
			Model & {mAP(\%)} & {mAP$_{50}$(\%)} & {mAP$_{75}$(\%)} \\
			\midrule
			Our model & 46.06 & 92.18 & 39.86 \\
			-w/o text & 35.64 & 81.68 & 29.10 \\
			-w/o LSRGAN & 34.82 & 79.54 & 28.67 \\
			\bottomrule
	\end{tabular}}
\end{table}
\subsection{Comparison of datasets}
\par To explore the performance of our model, we manually selected high-quality images. To evaluate the performance of our model, we created the HIP102 multi-modal data set manually, selecting high-quality images. HIP102 surpasses other multi-modal datasets in image resolution, quality, pest variety, and robustness. As demonstrated in Figure \ref{img11}, our model achieves higher accuracy, mAP, mAP$_{50}$ and mAP$_{75}$ on HIP102 compared to other datasets. This suggests that the IP102-based multi-modal dataset could still be improved in these areas.
Given that pests inhabit diverse environments and vary widely in size and species, we further examined the model’s ability to identify multiple targets at various scales within complex scenes. We constructed the MTIP102 multi-modal dataset to simulate these real-world conditions using the ACIE data enhancement method proposed in this study. MTIP102 includes images with multiple pest targets of different types and sizes.
\par In comparison to other single-target multi-modal datasets, MTIP102 achieves slightly lower scores in terms of accuracy, mAP, mAP$_{50}$, and mAP$_{75}$. However, it still outperforms the original IP102 unimodal dataset in several metrics. The decrease in performance can be attributed to the increased complexity of multi-target images, where the variations in pest types and scales pose greater challenges for feature learning. Nevertheless, MTIP102 closely simulates the natural environments of pests, enhancing the model’s robustness under challenging conditions.
Other datasets, such as STIP102 and CTIP102, were constructed in addition to HIP102 and MTIP102. While these datasets differ in complexity and target representation, they allow further refinement. This section highlights the need for dataset enhancement to improve model performance, especially in complex, real-world pest identification scenarios.
\subsection{Visualization of Regions of Interest } 
\begin{figure}[htp]
	\centering
	\includegraphics[height=5.5cm]{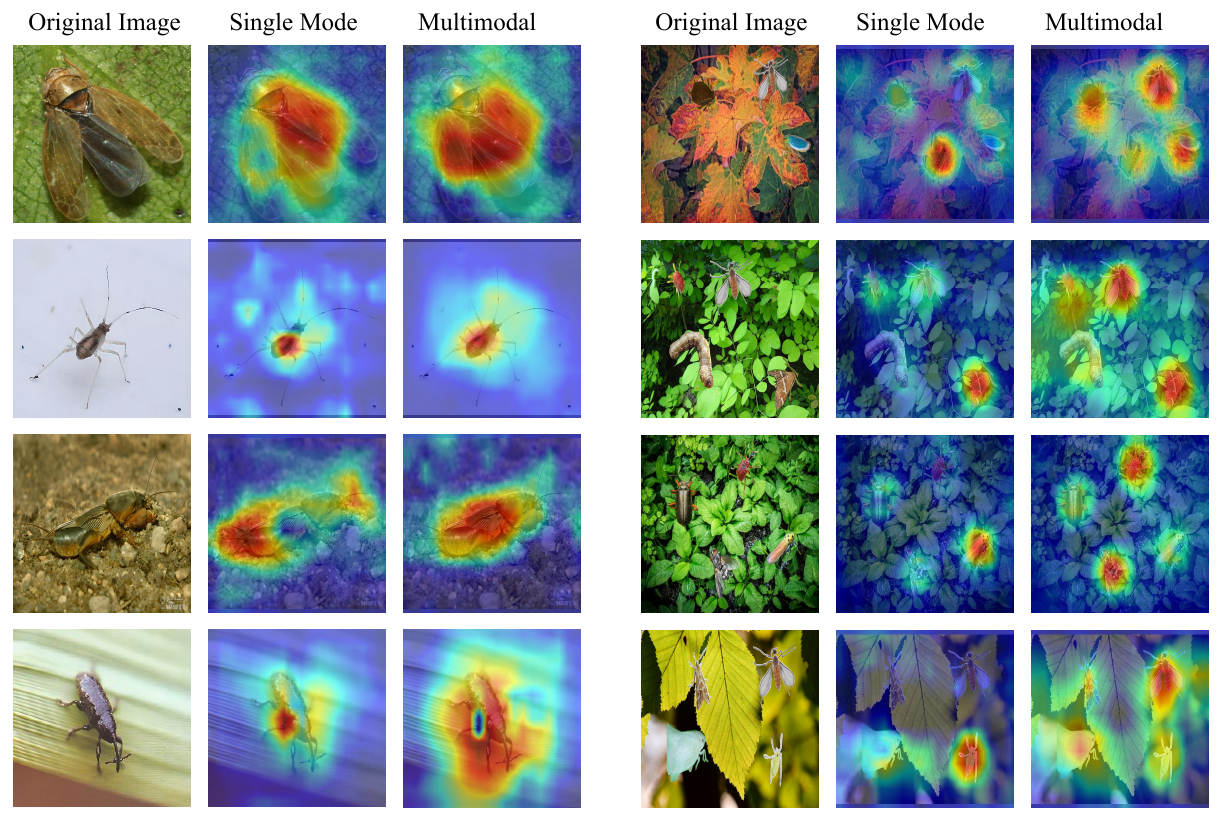}
	\caption{Grad-cam Visualization Results. (Single modal images are generated using w/o text, while multi-modal images are generated using MSFNet-CPD).}
	\label{img12}
\end{figure}

In this section, Grad-cam methods\cite{a31,a32} are employed to conduct an interpretable study of the model’s feature points of interest. The most critical locations in the Grad-cam plot correspond to the targets, which also serve as the model’s regions of interest (ROI). As depicted in Figure \ref{img12}, the Grad-cam plot of multi-modal images predominantly focuses on the primary areas of pests, reducing redundant coverage compared to unimodal images. This observation underscores the inadequacy of semantic information in unimodal visual features, highlighting the enhanced robustness of multi-modal features derived from the fusion of text and visual information.
\subsection{Discussion}
Although image-based methods have been widely used for disease or pest detection, there are still limitations in that images may not provide detailed information about pest species and characteristics\cite{a7,a32,a33}. The ITF structure proposed in this study solves this problem by processing text description information in images for feature extraction and fusion. The  MSFNet-CPD model has an advantage when image and text data are available, enabling more accurate and comprehensive pest classification and detection. Experiments have demonstrated that complex text descriptions can improve the detection rate, and future research should focus on the preprocessing of natural language descriptions and validate the model's performance in practical applications. In addition, this study proposes the ACIE data enhancement method, which can be applied to single-target detection with fewer data and extend the dataset, thus reducing the time for data labeling and collection.
\section{Conclusion}
This study addresses the challenges in pest detection, including complex and dynamic backgrounds and the limitations of single-mode detection. We propose a deep learning network, MSFNet-CPD, which utilizes multi-scale cross-modal fusion network of image and text data to detect and classify 102 pest species using the IP102 dataset. Experimental results show that integrating a super-resolution reconstruction algorithm with the multi-scale cross-modal fusion strategy significantly enhances the model's performance. Specifically, MSFNet-CPD achieves a detection precision of 82.15\% and a mAP score of 46.06\% when processing both image and text inputs simultaneously.
\par Ablation studies demonstrate that our approach not only overcomes the limitations of image-only features but also benefits from higher-quality data, improving network performance. Furthermore, we expect even better performance under real-world conditions by incorporating multi-target datasets with diverse pest species and sizes, enhanced through the ACIE data augmentation algorithm.
\par Overall, our MSFNet-CPD network for cross-modal, multi-scale fusion shows strong potential for pest identification in complex environments, with promising applications for a wide range of agricultural tasks in the future.
\bibliographystyle{IEEEtran}
\bibliography{ref}
\end{document}